\documentclass[journal]{IEEEtran}
\usepackage{cite}
\usepackage{amsmath,amssymb,amsfonts}
\usepackage{algorithmic}
\usepackage{graphicx}
\usepackage{textcomp}
\usepackage{array}
\usepackage{booktabs}
\usepackage{multirow}
\usepackage{subfig}
\usepackage{wrapfig}

\begin{document}

\title{RangeSeg: Range-Aware Real Time Segmentation of 3D LiDAR Point Clouds}

\author{Tzu-Hsuan Chen, and Tian Sheuan Chang, \IEEEmembership{Senior Member, IEEE}

\thanks{This work was supported by the Ministry of Science and Technology, Taiwan, under Grant 109-2634-F-009 -022 and 109-2639-E-009-001. The authors are with the Department of Electronics Engineering, National Chiao Tung University, Hsinchu 30010, Taiwan (e-mail: jesse.ee06g@nctu.edu.tw, tschang@mail.nctu.edu.tw). \\
© 2021 IEEE.  Personal use of this material is permitted.  Permission from IEEE must be obtained for all other uses, in any current or future media, including reprinting/republishing this material for advertising or promotional purposes, creating new collective works, for resale or redistribution to servers or lists, or reuse of any copyrighted component of this work in other works.
Cite this as \\
T. -H. Chen and T. S. Chang, "RangeSeg: Range-Aware Real Time Segmentation of 3D LiDAR Point Clouds," in IEEE Transactions on Intelligent Vehicles, vol. 7, no. 1, pp. 93-101, March 2022, doi: 10.1109/TIV.2021.3085827.
}}

\maketitle
\begin{abstract}
Semantic outdoor scene understanding based on 3D LiDAR point clouds is a challenging task for autonomous driving due to the sparse and irregular data structure.  This paper takes advantages of the uneven range distribution of different LiDAR laser beams to propose a range aware instance segmentation network, RangeSeg.  RangeSeg uses a shared encoder backbone with two range dependent decoders.  A heavy decoder only computes top of a range image where the far and small objects locate to improve small object detection accuracy, and a light decoder computes whole range image for low computational cost.  The results are further clustered by the DBSCAN method with a resolution weighted distance function to get instance-level segmentation results.  Experiments on the KITTI dataset show that RangeSeg outperforms the state-of-the-art semantic segmentation methods with enormous speedup and improves the instance-level segmentation performance on small and far objects.  The whole RangeSeg pipeline meets the real time requirement on NVIDIA\textsuperscript{\textregistered} JETSON AGX Xavier with 19 frames per second in average.

\end{abstract}



\section{Introduction}
The semantic scene understanding from 3D LiDAR point clouds is one of the fundamental blocks to provide robust and real time 3D object detectors for the autonomous driving.  LiDAR sensors provide range measurements by sampling a specific location with spanning vertical and horizontal angular resolutions, which is different from the 3D point clouds sampled densely and uniformly on all sides used in indoor scenes.  In addition, LiDAR sensors are robust under almost all light conditions or the foggy weather.  As a result, 3D LiDAR point clouds attract the significant research attention recently. 

The major difficulty in processing LiDAR data is that the sensors provide non-Euclidean data in the form of point clouds with about 100k points around 360\textdegree, which poses a great challenge for object detection and segmentation tasks and thus needs the high computational cost.  For the 3D object detection and the 3D semantic segmentation, the segmentation task gives the dense predictions of the scene understanding.  The previous work such as SqueezeSeg \cite{squeezeseg} proposed a light weight convolutional neural network (CNN) backbone with the conditional random field (CRF) \cite{deeplab} to get the real time performance, but it still leaves room for accuracy improvements.  Higher accuracy demands higher computational cost, which hurts real time performance.

\begin{figure}[t]
    \centering
    \includegraphics[height=!,width=\linewidth,keepaspectratio=true]%
    {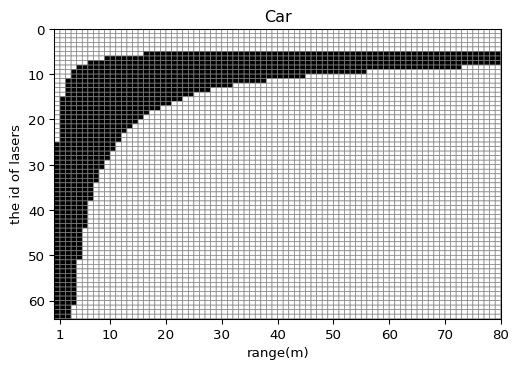}
    \caption{Range distributions of the detected cars (denoted by black) for different ID of lasers in LiDAR.  The simulation setting is based on Velodyne HDL-64E S3.}
    \label{fig:laserid}
\end{figure}

To achieve high accuracy within a real time constraint, this paper proposes a instance-level segmentation, denoted as RangeSeg.  RangeSeg exploits the uneven range distribution of 3D LiDAR point clouds as shown in Fig.~\ref{fig:laserid} for the autonomous driving. It has far and small objects at the top of the range image, and large and near objects at the bottom of the range image.  This algorithm achieves high accuracy for these far and small objects by adopting a heavy decoder only on the top of the image, and meets the real time demand by adopting a light decoder on the whole image with a shared backbone encoder.  The semantic segmentation results are further clustered as instances by the density-based spatial clustering and applications with noise (DBSCAN) \cite{dbscan} method based on a resolution weighted distance.  The result shows that the proposed method can improve the detection on far and small objects with the real time execution performance on NVIDIA\textsuperscript{\textregistered} JETSON AGX Xavier.

The rest of the paper is organized as follows. Section II first introduces the related works. Section III presents our approach. Section IV shows the experimental results and comparisons with other methods. Finally, we conclude this paper in Section V.

\section{Related Works}
\subsection{Data representation of 3D LiDAR point clouds}
3D LiDAR point clouds have an unstructured data format.  To tackle such unstructured data for 3D outdoor scene understandings, one approach is to transform point clouds into a structured format to utilize standard convolutional operations.  The other approach is to define a new operation directly on unstructured data.  Current data transformations are mainly divided into two types: 3D voxel grids or 2D projections.  The 3D voxel grid method transforms the data into a regular space of 3D grid such that the following 3D convolution operations can be applied to extract the high order of feature representations.  However, the 3D voxel grids are sparse since the point clouds are inherently sparse, which wastes lots of computations on unnecessary grids.  The 2D projections such as the birds' eye view (BEV) and the range image encoding are much more compact.  BEV is sparse while preserves the size of objects.  The range image encoding is dense but distorts objects.  The novel graph-based neural networks can directly apply on point clouds, but the recent approaches only apply on the 3D indoor scene understanding, in which the point clouds are uniform sampled on surface with about 1k points.
\subsection{3D Object Detection of Point Clouds}
VoxelNet \cite{voxelnet} encodes point clouds into hand-crafted 3D voxel grids by a voxel feature encoding layer to extract high order of features.  They use 3D CNN layers to aggregate the voxel-wise features with expanded receptive fields.  However, the 3D CNN has high computational cost even for such sparse representation. Its real time performance is limited to four frames per second (fps).  PIXOR \cite{pixor} uses a 2D birds' eye view representation to build a real time pipeline, but fails to deliver a good performance on small objects.  PointRCNN \cite{pointrcnn} is the first two-stage 3D detector that only uses 3D LiDAR point clouds, which uses PointNet++ \cite{pointnet++} on the unstructured data to get the preliminary bounding boxes and applies a simple multi-layer neural network to get a final prediction.  This approach gets a good performance on small objects such as pedestrians and cyclists, but the two-stage pipeline makes it unsuitable for real time applications.

\subsection{3D Semantic Segmentation of Point Clouds}
SqueezeSeg \cite{squeezeseg}, PointSeg \cite{pointseg} and SqueezeSegV2 \cite{squeezesegv2} all use a light weight SqueezeNet \cite{squeezenet} as their backbone with range images as input for the semantic segmentation task.  They use different post-processing methods to improve the accuracy.  SqueezeSeg and SqueezeSegV2 use a recurrent CRF module \cite{deeplab} to reduce the blurry boundaries.  SqueezeSegV2 additional tackles the inherent problems of missing points in range images with a context aggregation module.  PointSeg uses a squeeze re-weighting layer \cite{squeezeexcitation} and an enlargement layer \cite{deeplab} to achieve a better performance.  RIU-Net \cite{riunet} directly uses U-Net \cite{unet} on range images with focal loss \cite{focal}.  These works get real time performance due to a light weight backbone but has a low accuracy.  Moreover, none has discussed the instance-level segmentation. 

\subsection{Graph Neural Networks of Point Clouds}
Pointnet \cite{pointnet} proposed to use an end-to-end pipeline to learn point-wise features directly from point clouds.  The follow-up work improves the performance by extracting local features \cite{pointnet++}.  Furthermore, DGCNN \cite{dgcnn} and PointCNN \cite{pointcnn} define a new convolution operation on point clouds.  They succeed in the 3D indoor scene understanding ($\sim$1k points).  However, the outdoor scene contains about 100k points, which will make the above network training demand high requirements of  the memory and the computation.

\subsection{3D Instance Segmentation of Point Clouds}
A novel paper \cite{zhang2020instance} proposed a pipeline with a  feature learning network and a stacked hourglass network for the instance segmentation in the outdoor LiDAR point clouds, which could help localize the small and far-away objects.  However, the high complexity of the model makes it hard to run in real time, where the elapsed time in TESLA V100 GPU was 300 ms.  

\section{RangeSeg Framework}
\begin{figure}[t]
    \centering
    \includegraphics[height=!,width=1\linewidth,keepaspectratio=true]%
    {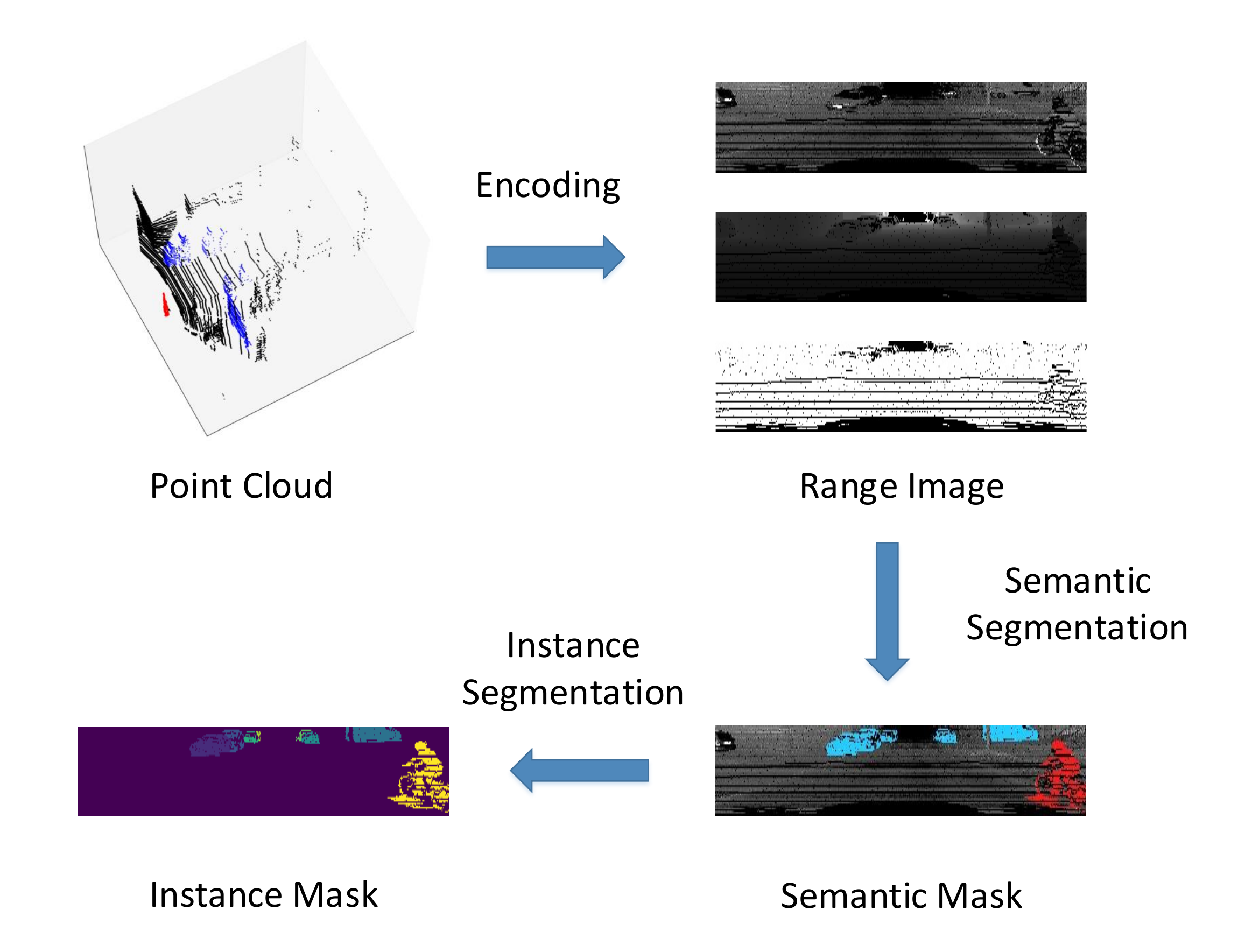}
    \caption{The proposed pipeline of RangeSeg.}
    \label{fig:pipeline}
\end{figure}
In this paper, we propose a real time pipeline, RangeSeg, that gets accurate instance-level segmentation results by exploiting 2D range image representations of LiDAR point clouds.  Our range-aware framework attains a fast and accurate semantic segmentation by a complex heavy decoder to predict far and small objects and a light decoder to reduce complexity of general predictions.  Next, we use DBSCAN with the proposed weighted distance function to get instance-level segmentation results.  An overview of the whole pipeline is shown in Fig.~\ref{fig:pipeline}.  In the following subsections, we will introduce our input representation, range-aware network architecture and how to use DBSCAN as a post-processing to get instance-level segmentation results.
\subsection{Input Representation}
3D point clouds are unstructured data while the standard neural networks perform discrete convolution operations on grids.  Thus, several methods have been proposed to encode point clouds into a suitable format.  In which, the 3D voxel grids are one possible solution.  However, 3D convolution operations are computational intensive, and the sparse voxel grids will lead to lots of unnecessary computations.  The 2D birds' eye view is another solution, but leads to information loss.  Instead, we use range images to represent the point clouds.  The range image is a 2D dense image-like data format without the information loss, and does not need hand-crafted parameters during the format conversion.

The point clouds are converted to range images as following. First, project the points onto a spherical coordinate system with the grid-based representation as
\begin{equation}
   \begin{aligned}
      \theta &= \arctan{\left(\frac{z}{\sqrt{x^2+y^2}}\right)} &,\hat{\theta} &= \left \lfloor{\frac{\theta}{\Delta\theta}} \right \rfloor \\\\
      \phi &= \arctan{\left(\frac{y}{x}\right)} &,\hat{\phi} &= \left \lfloor{\frac{\phi}{\Delta\phi}} \right \rfloor\\
   \end{aligned}
   \label{eq:encoding}
\end{equation}
where $\theta$ and $\phi$ are an azimuth angle and an elevation angle respectively. 
$\Delta\theta$ and $\Delta\phi$ are resolutions for the discretization and $(\hat{\theta},\hat{\phi})$ denotes the position of 2D spherical grids.  Applying (\ref{eq:encoding}) on each point, we can get a 3D $H\times W\times K$ tensor.  In this paper, we use the KITTI dataset \cite{kitti3d,kittiraw} collected by Velodyne HDL-64E S3, which has 64 laser beams, ($H=64$).  Also, the horizontal angular resolution is 0.1728\textdegree\ and the annotations are only available in the 90\textdegree\ front view, ($W=512$).  K is the number of features, which is 3 in this paper, ($K=3$), including intensity, range measurements and occupancy.  The occupancy channel indicates whether the grids contain points.  The visualization of a range image representation is shown in Fig.~\ref{fig:pipeline}.
\subsection{Range Distribution on Range Images}
In order to get 360\textdegree\ view of scenes, the LiDAR sensors are mostly placed on the top of vehicles.  As a consequence, the vertical view is asymmetric, where only few laser beams are emitted to the upper part of a scene.  The lower position lasers can only sample the objects in a much shorter range.  Fig.~\ref{fig:laserid} shows the laser ID that can detect "car" at different ranges.  As shown in the figure, only the top part of lasers can detect objects in the whole range, especially if they are far away.  Besides, the far objects only occupy few pixels on range images due to the lower density of points associated to objects at larger distances.  Based on the observation, we propose the range-aware framework that uses different decoders for different parts of the range image to get higher accuracy and speedup as well.
\subsection{Network Architecture}
\begin{figure*}[t]
    \centering
    \includegraphics[height=!,width=\linewidth,keepaspectratio=true]%
    {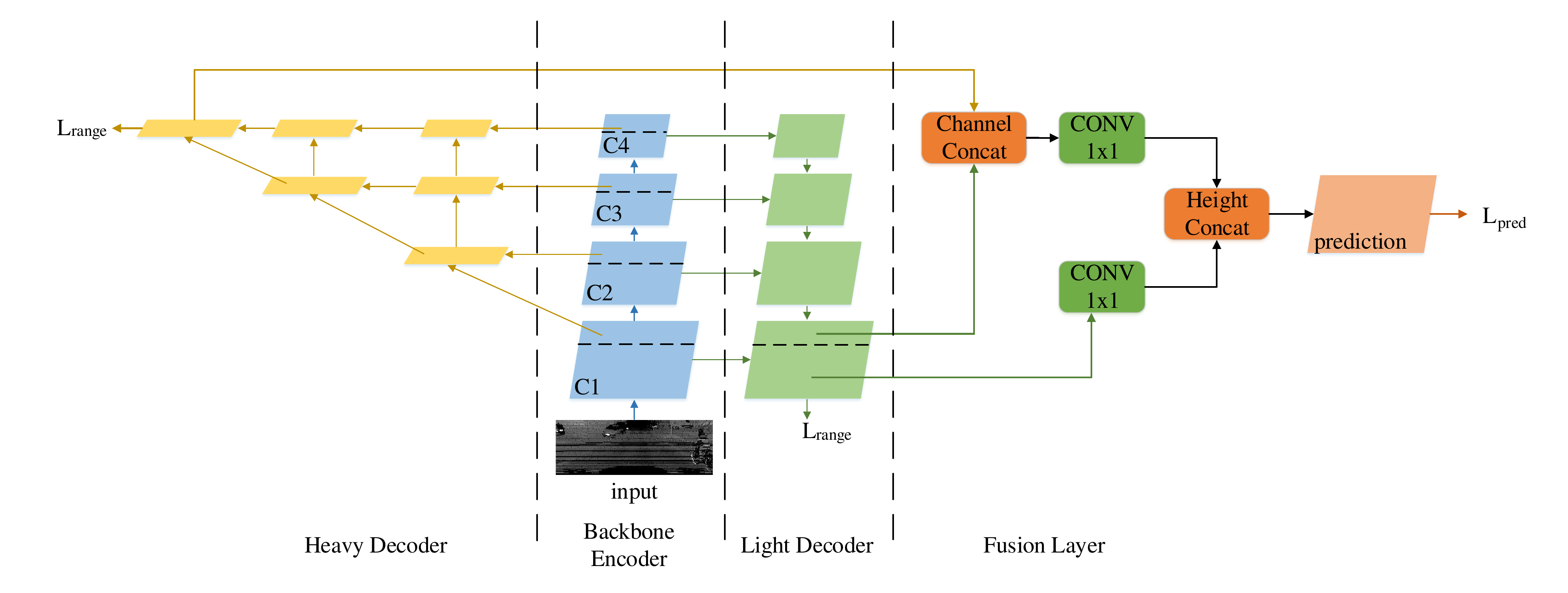}
    \caption{The architecture of the proposed range-aware network.}
    \label{fig:scaleaware-arch}
\end{figure*}
\begin{figure}[t]
    \centering
    \subfloat[ResNet18]{
        \includegraphics[height=!,width=0.5\linewidth,keepaspectratio=true]%
    {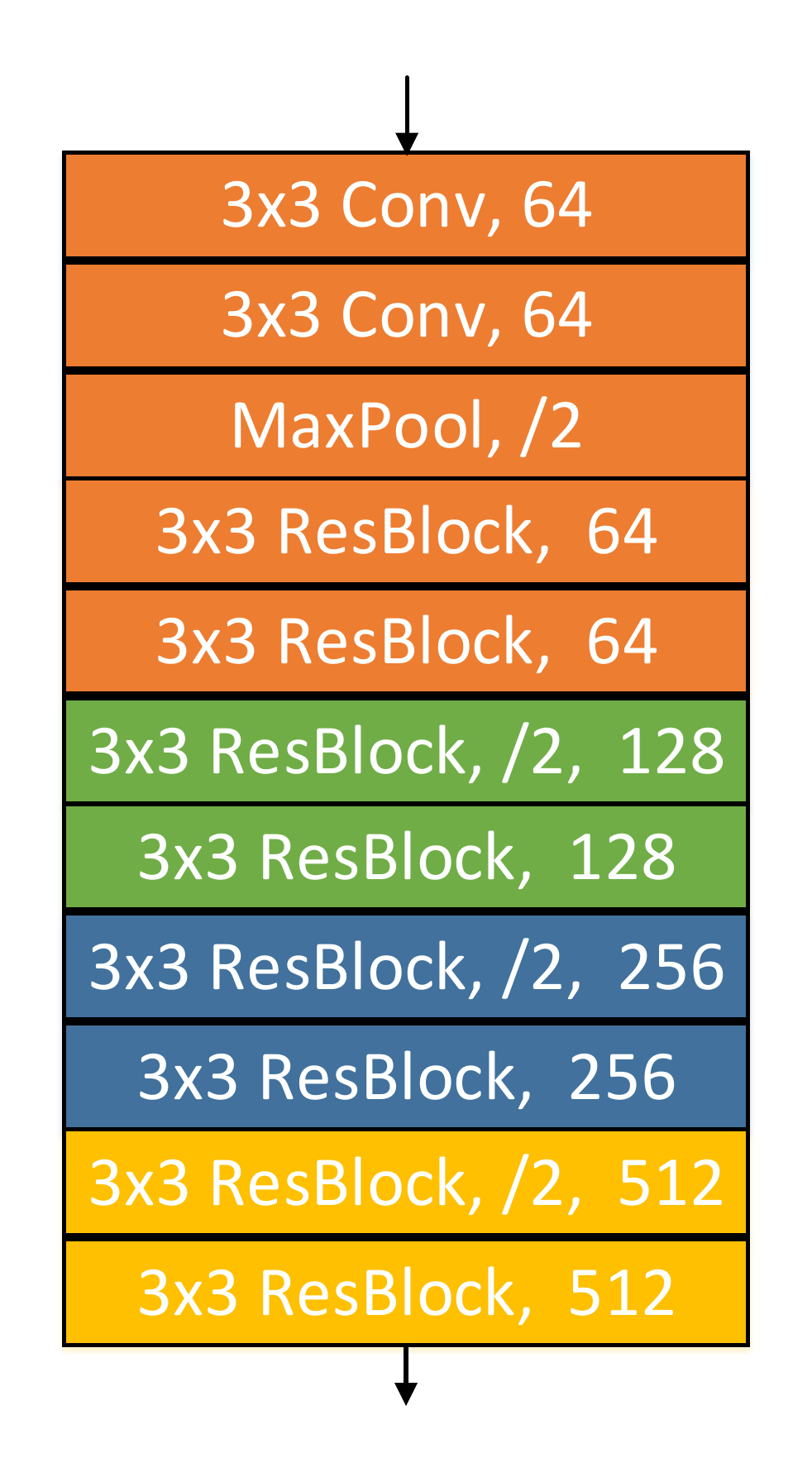}
    \label{fig:resnet18-arch}
    }
    \subfloat[LaserNet]{
        \includegraphics[height=!,width=0.5\linewidth,keepaspectratio=true]%
    {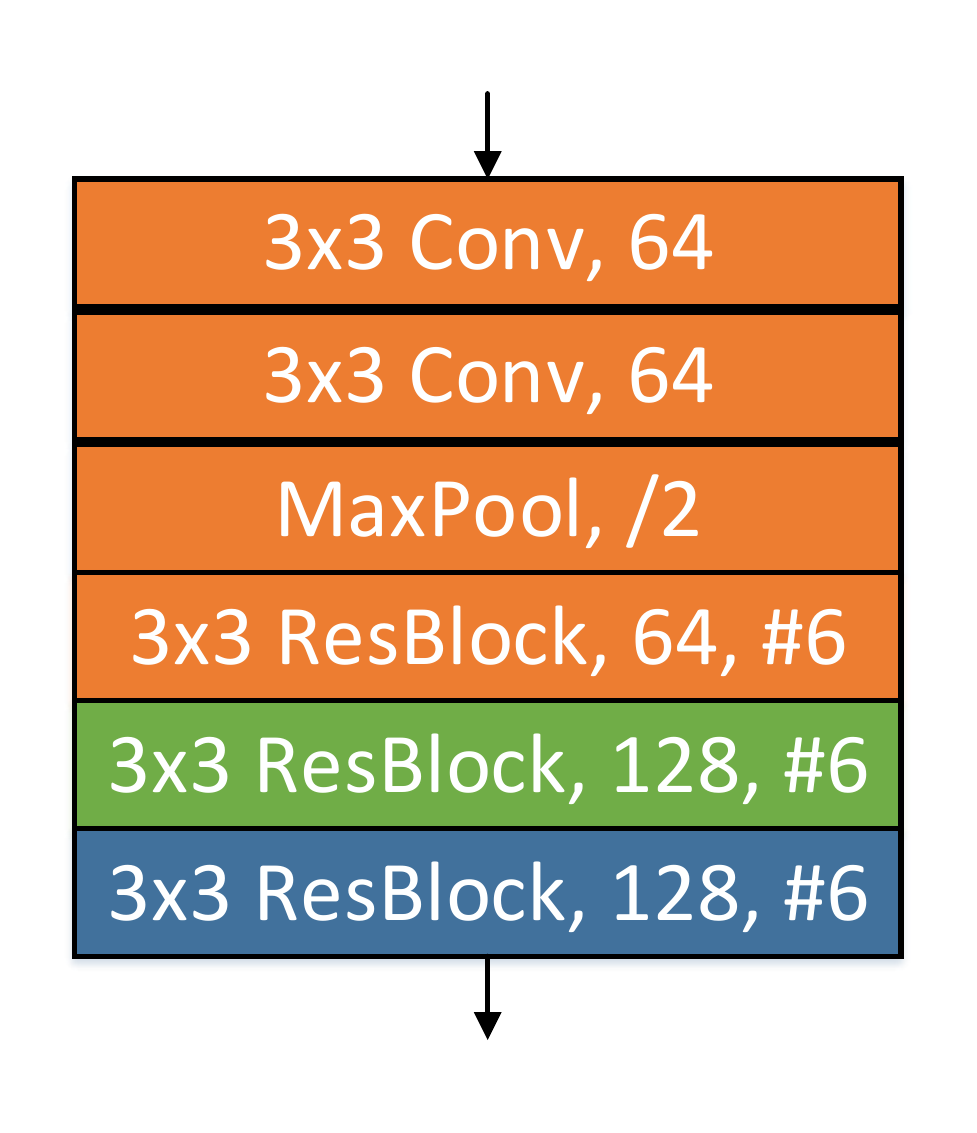}
    \label{fig:lasernet-arch}
    }
    \caption{The architecture of the modified ResNet18 and LaserNet, where \# denotes the number of blocks.}
    \label{fig:backbone-arch}
\end{figure}
RangeSeg as shown in Fig.~\ref{fig:scaleaware-arch} is a fully-convolutional network with one backbone encoder, two decoders, a fusion layer and a simple post-processing for final instance-level segmentation prediction.  The outputs of network are the same size as inputs to get higher accuracy segmentation results.

For the encoder backbone, this paper uses two different backbones for testing.  The first one is the modified version of ResNet-18 \cite{resnet}.  The original ResNet-18 performs two down-sampling steps at the beginning of the network, which are removed in this paper to enable computations on original resolution feature maps for better accuracy, as shown in Fig.~\ref{fig:backbone-arch} (a).  The other one is based on LaserNet \cite{lasernet} that performs 3D object detection on range images.  In which, the residual blocks are also used to get better performances with deeper layers and fewer channels.  Although the LaserNet backbone is much deeper, its parameters of kernels are fewer due to the fewer channels as shown in Fig.~\ref{fig:backbone-arch} (b).

For the target KITTI dataset, the input representation is $64\times 512\times 3$.  Thus, we only perform the stride and downsample operations on the vertical dimension to minimize information loss.

\begin{figure}[t]
    \centering
    \subfloat[The light decoder]{
        \includegraphics[height=!,width=\linewidth,keepaspectratio=true]%
    {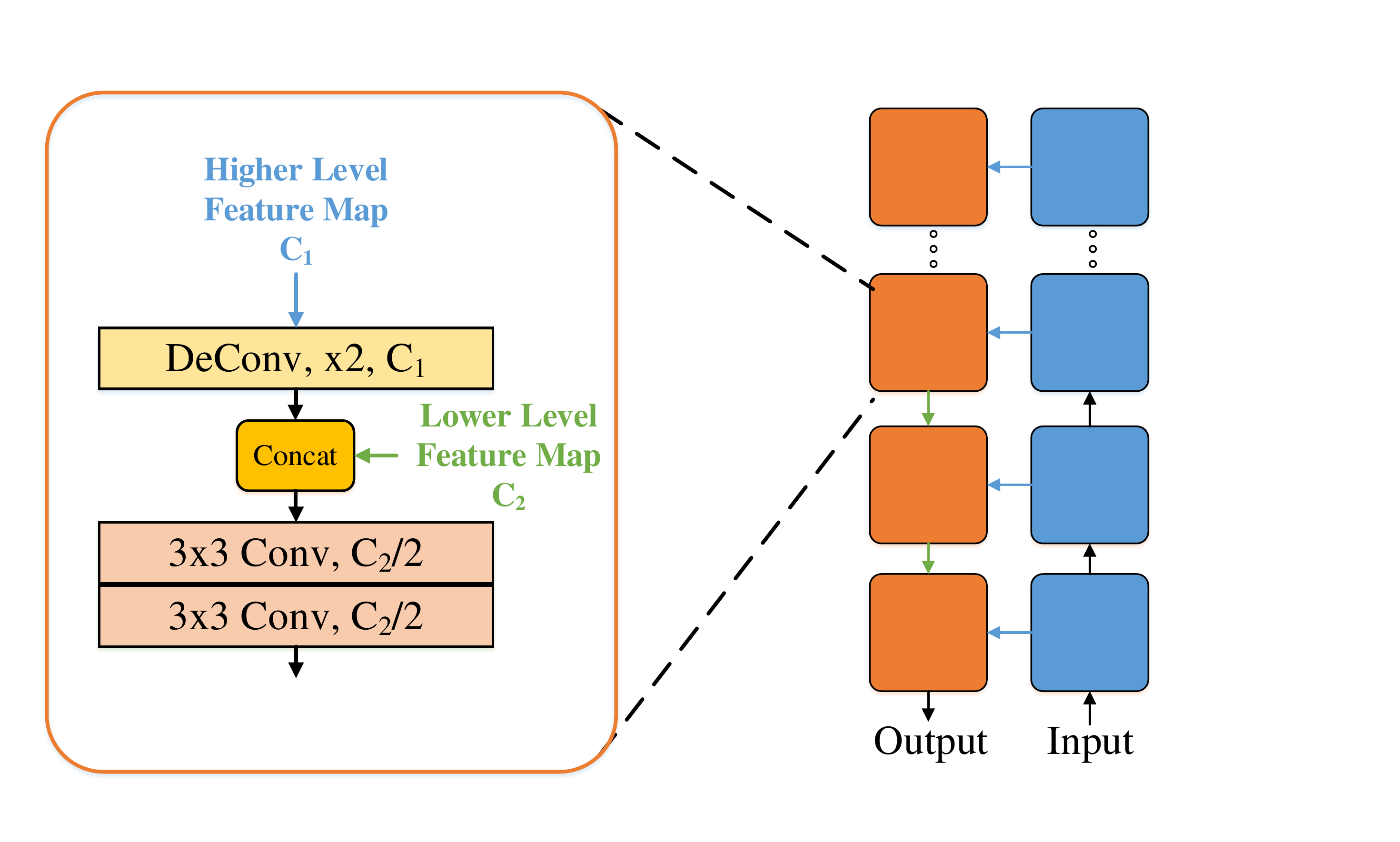}
    } \\
    \subfloat[The heavy decoder]{
        \includegraphics[height=!,width=\linewidth,keepaspectratio=true]%
    {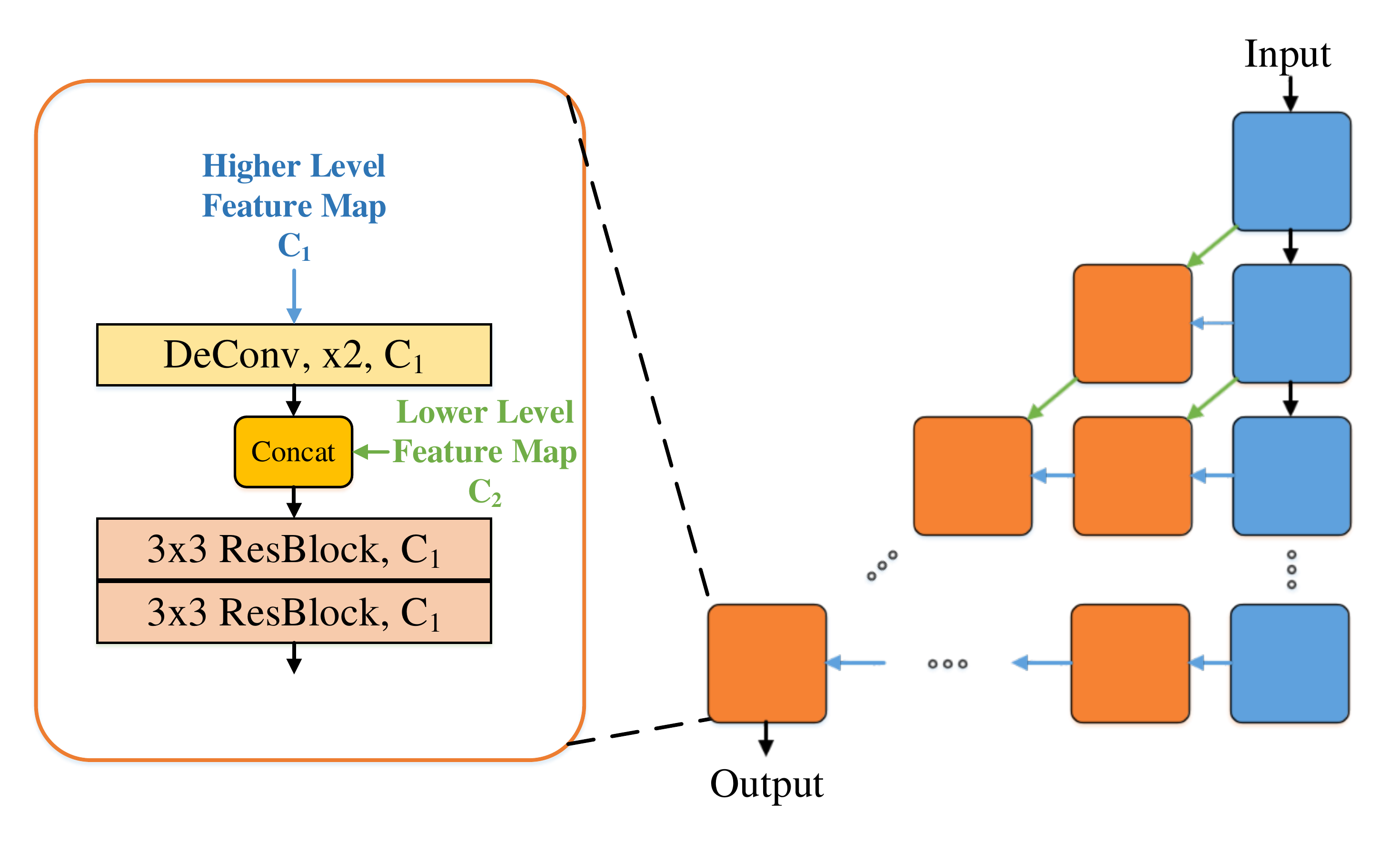}
    }
    \caption{The architecture of the heavy and light decoders.  The blue block is the feature extractor and the orange block is the up-sample module.  The black arrow denotes the normal path of the feature extractor.  The blue one denotes the input of the higher level feature maps.  The green one denotes the input of the lower level feature maps.  The details of up-sample module are shown in the left part.}
    \label{fig:decoder-arch}
\end{figure}

The range-aware decoders use the same feature maps from the backbone network for the heavy and light decoders to exploit the different range distribution of range images.  The \textit{heavy decoder} only predicts the results for top of images, where the small and far objects locates and needs the deeper network aggregation for accurate predictions.  The heavy decoder uses the `deep' skip connections inspired by DLA \cite{dla}, where the high level feature maps will be upsampled and aggregated with lower level ones repeatedly.  In this paper, instead of using the tree-structured DLA, a much dense concatenation is used as shown in Fig.~\ref{fig:decoder-arch} (b) since this decoder only processes the top rows of the range maps.  The \textit{light decoder} predicts the results of whole images with low computational cost.  The light decoder uses the `shallow' skip connections like U-Net \cite{unet} that only concatenates the feature maps once as shown in Fig.~\ref{fig:decoder-arch} (a).  It has low computational cost while preserves the information of different range features.  

The predictions from the light and heavy decoders are fused together for final results.  However, these two predictions have different spatial sizes, which prohibits a direct fusion.  For a smooth fusion, the whole output of the heavy decoder is concatenated with the top same size output from the light decoder.  Then the concatenated result changes its channel numbers by 1x1 convolutions to match the channel numbers of the bottom part from the light decoder, as shown in Fig.~\ref{fig:scaleaware-arch}.  Finally, these two parts are concatenated along the height dimension for the final results.  In order to get the predictions on the same resolution, a transpose convolution layer is applied after the fusion layer.  In our experiments, we apply the heavy decoder on the top 16 rows of range images based on the observations of the KITTI dataset.
\subsection{Training Strategy}
RangeSeg uses the common multi-class cross entropy loss and Lov\'{a}sz-softmax loss \cite{lovasz} to train the network.  Cross-entropy loss is used for the pixel-wise classification loss on the classification output $p$ with the target $y$.  If $k$ denotes the channels of the input images, cross-entropy loss is defined as
\begin{equation}
   Loss_{xent}(\boldsymbol{p},\boldsymbol{y})=\frac{1}{HWK}\sum_{i=0}^{H}\sum_{j=0}^{W}\sum_{k=0}^{K}-y_{i,j,k}\log  p_{i,j,k}
   \label{eq:lossxent}
\end{equation}
However, cross-entropy loss is not directly related to intersection over union (IoU).  Therefore, Lov\'{a}sz-softmax loss, which is a Lov\'{a}sz extension of the Jaccard index, is applied to regularize the network.  If $c$ denotes the class, Lov\'{a}sz-softmax loss is defined as
\begin{equation}
   m_{i,j}(\boldsymbol{p},\boldsymbol{y},c) = 
   \begin{cases}
       1-p_{i,j}(c),& \text{if }c=y_{i,j} \\
       p_{i,j}(c),& \text{otherwise}
   \end{cases}
   \label{eq:pixelerror}
\end{equation}
\begin{equation}
   L_{Lov\acute{a}sz}(\boldsymbol{p},\boldsymbol{y})=\frac{1}{C}\sum_{c=0}^{C}\overline{\Delta_{J_c}}\left (\boldsymbol{m}\left (\boldsymbol{p},\boldsymbol{y},c\right )\right )
   \label{eq:lovasz}
\end{equation}
where $\overline{\Delta_{J_c}}$ is the surrogate function of Jaccard loss ($\Delta_{J_c}$).  Lov\'{a}sz-softmax loss is IoU-aware and helps solve the data imbalance.  Therefore, the loss function, $L(\boldsymbol{p},\boldsymbol{y})$, is defined as
\begin{equation}
   L(\boldsymbol{p},\boldsymbol{y}) = L_{xent}(\boldsymbol{p},\boldsymbol{y})+\lambda_{Lov\acute{a}sz}L_{Lov\acute{a}sz}(\boldsymbol{p},\boldsymbol{y})
   \label{eq:lossoutput}
\end{equation}
with the Lov\'{a}sz weighted term $\lambda_{lov\acute{a}sz}$.  Therefore, the loss of the prediction part from the fusion layer as in Fig.~\ref{fig:scaleaware-arch} is defined as
\begin{equation}
   L_{pred} = L(\boldsymbol{p}_{pred},\boldsymbol{y}_{pred})
   \label{eq:predloss}
\end{equation}
The range-aware decoder loss combines the results from heavy and light decoders directly as a regularization term instead of computing loss on results of the fusion layer alone, which is defined as
\begin{equation}
   L_{range} = L(\boldsymbol{p}_{light},\boldsymbol{y}_{light})+L(\boldsymbol{p}_{heavy},\boldsymbol{y}_{heavy})
   \label{eq:scaleloss}
\end{equation}
The total network loss is defined as

\begin{equation}
   L_{total} = L_{pred}+\lambda_{range}L_{range}
   \label{eq:totalloss}
\end{equation}
with a range-aware weighted term $\lambda_{range}$.  In our experiments, we set both $\lambda_{lov\acute{a}sz}$ and $\lambda_{range}$ as 1.

We use the super convergence strategy\cite{superconvergence} as our learning rate scheduler, 
which has higher and dynamic learning rate for fast convergence.

\subsection{Resolution Weighted Distance Function for Instance Segmentation}
\begin{table*}[t]
    \centering
    \begin{tabular}{ c  >{\centering\arraybackslash}m{1.5cm}  >{\centering\arraybackslash}m{1.5cm}  >{\centering\arraybackslash}m{1.5cm} >{\centering\arraybackslash}m{1.5cm} >{\centering\arraybackslash}m{2cm} >{\centering\arraybackslash}m{2cm} c }
       \toprule
       \multirow{3}{*}{Architecture} & \multicolumn{6}{c}{IoU(\%)} & \multirow{3}{*}{FPS}\\
       \cline{2-7}
       & \multirow{2}{*}{Car} & \multirow{2}{*}{Pedestrian} & \multirow{2}{*}{Cyclist} & Overall & Top 16 Rows & Lower 48 Rows &  \\
       & & & & Mean & Mean & Mean & \\
       \toprule
       UNet & 75.9 & 66.7 & 47.4 & 63.3 & 62.9 & 63.6 & 249.8 \\
       \midrule
       ResNet18-UNet & 74.8 & 65.3 & 46.0 & 62.0 & 62.4 & 61.3 & \textbf{122.9} \\
       ResNet34-UNet & 76.0 & 67.2 & \textbf{53.5} & 65.6 & 65.2 & \textbf{65.6} & 76.6 \\
       Ours(ResNet18) & \textbf{76.7} & \textbf{68.5} & 52.5 & \textbf{65.9} & \textbf{67.3} & 64.1 & 89.2 \\
       \midrule
       LaserNet-DLA & 77.3 & 69.1 & 51.7 & 66.0 & 67.6 & 64.0 & 77.0 \\
       Ours(LaserNet) & \textbf{77.8} & \textbf{70.2} & \textbf{56.2} & \textbf{67.9} & \textbf{69.4} & \textbf{66.0} & \textbf{109.1} \\
       \bottomrule
    \end{tabular}
    \caption[Result comparison on the KITTI 3D object detection benchmark.]{Result comparison on the KITTI 3D object detection benchmark.  ResNet-UNet uses ResNet as the encoder backbone with the up-sample layers as UNet.  LaserNet-DLA use LaserNet as the encoder backbone with the fully DLA-like feature aggregation.}
    \label{table:kitti3d}
 \end{table*}


To the best of authors' knowledge, none works have used 3D information to segment instances.  Unlike objects in the 2D RGB images, the 3D objects will not be overlapped in the 3D space domain, which will make it much easier to segment different objects.  Therefore, this paper uses the density-based spatial clustering applications with noise (DBSCAN) \cite{dbscan} for instance clustering, which does not require a pre-defined number of clustering.  The clusters are defined by their density. In this paper, DBSCAN takes the points labeled as objects after semantic segmentation as input.  Then, the clustering process is applied once for all the objects to save computations for background points and multiple iterations. For the DBSCAN distance function, instead of directly using vanilla distance function, we propose a weighted distance function to deal with the resolution differences of the LiDAR data.  The resolution of the vertical dimension is twice fewer than the horizontal one, which leads to sparser vertical values compared with the horizontal ones.  Therefore, a resolution weighted distance function is defined as
\begin{equation}
   Dis. = \sqrt{2 \left ( x_2-x_1\right )^2 + 2\left ( y_2-y_1\right )^2 + \frac{\left ( z_2-z_1\right )^2}{2}}
   \label{eq:distance}
\end{equation}
This distance function scales up the coordinates X and Y as well instead of scaling down the coordinate Z alone since scaling down the coordinate z alone cannot help segmentation if the distance is dominated by the horizontal one.

\begin{figure*}
    \centering
    \includegraphics[clip, trim=0.5cm 0.5cm 0.5cm 7cm, width=\linewidth,keepaspectratio=true]%
    {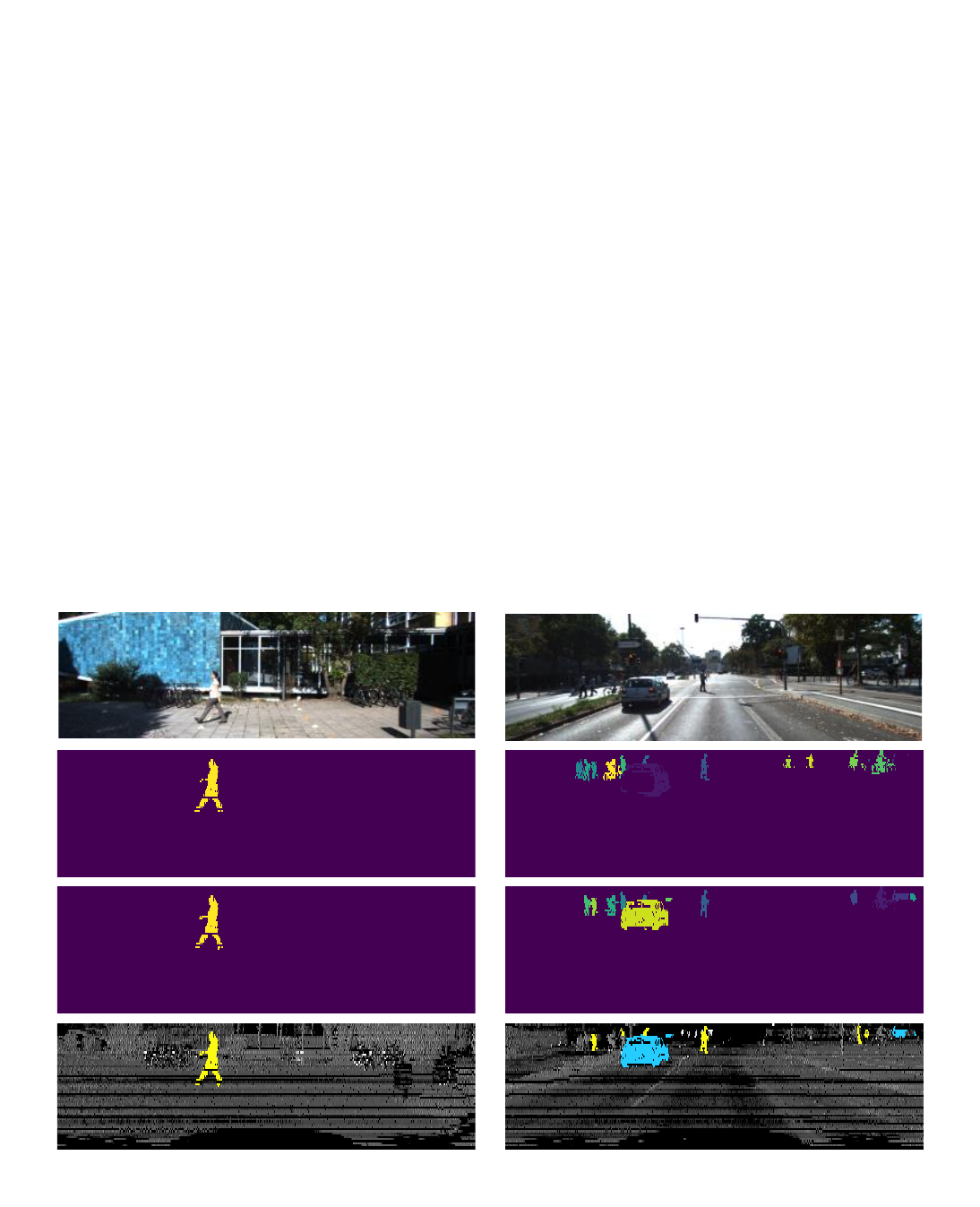}
    \caption{The visualization results of instance segmentation for different scenes.  For each scene, the images from top to bottom are raw images, ground truth, prediction by RangeSeg, prediction by RIU-Net. Though RIU-Net has good good real time performance and good accuracy for simple scenes, its false negative results become obvious for small or far objects when compared to the proposed approach. }
    \label{fig:instance-vis}
\end{figure*}
\section{Experimental Results}

We evaluate our model on the KITTI dataset and empirically showcase the strengths and weaknesses of the proposed approach.  First, we compare the vanilla segmentation frameworks with our range-aware framework using different backbones on the KITTI 3D object detection benchmark \cite{kitti3d}.  We show that our RangeSeg outperforms on accuracy and inference speed.  Second, the comparison of different distance functions on DBSCAN shows that the proposed distance function helps improve the accuracy.  Third, we compare RangeSeg with related works on the KITTI raw data, where the accuracy gets lots of improvement with the same inference speed.  Fourth, we implement RangeSeg on NVIIDA\textsuperscript{\textregistered} JETSON AGX Xavier, which gets real time performance even on such small embedded system.  Finally, the experiment results on synthetic foggy KITTI data \cite{foggykitti} show that our approach is still robust in the foggy weather.  The evaluation metric used in this paper is mean class IoU (mIoU).
\subsection{KITTI 3D Object Detection Benchmark}
\paragraph{Implementation Details}

We encode the front 90\textdegree\ into a $64\times 512\times 3$ tensor with 3 features: intensity, range and occupancy.  The value range of intensity is already within [0,1], and occupancy is either 0 or 1.  Therefore, we normalize the range features to be within [0, 1].  In addition, the random horizontal flip with probability 0.5 is applied as the augmentation.  For ground truth, we treat the points in the bounding boxes as the objects while the remains are background due to the limitations of the KITTI annotations.

\paragraph{Range-Aware Framework} 

Table~\ref{table:kitti3d} summarizes the comparisons between the vanilla segmentation frameworks and range-aware framework.  For ResNet based backbone, the range-aware ResNet18 leads the ResNet18-UNet by 3.9\%.  The improvements are more significant on small objects such as pedestrians and cyclists by 3.2\% and 6.5\%, respectively.  The range-aware ResNet18 even outperforms ResNet34-UNet by 0.3\%.  For LaserNet based backbone, the range-aware LaserNet leads LaserNet-DLA by 1.9\%.  Moreover, the results on top 16 rows of range images show that RangeSeg helps improve detection of the far and small objects since the far objects only lie in the top of images.

\paragraph{Inference Speed}
Table~\ref{table:kitti3d} shows the inference speed on the Nvidia TITAN Xp.  The range-aware ResNet18 improves 3.9\% than ResNet18-UNet with only 27\% fps loss, while improves 0.3\% than ResNet34-UNet with extra 7\% speedup.  Range-aware LaserNet improves 1.9\% of mIoU than LaserNet-DLA with extra 42\% speedup since the heavy decoder helps improvements on mIoU with computation overhead, but the low complexity light decoder helps overcome the problems.
\paragraph{Ablation Study}
\begin{table}[t]
    \centering
    \begin{tabular}{ c >{\centering\arraybackslash}m{0.7cm} >{\centering\arraybackslash}m{0.7cm} c c}
        \toprule
        Loss & Norm. & Aug. & Range & mIoU(\%) \\
        \midrule
        xent & - & - & - & 45.8 \\
        focal & - & - & - & 20.6 \\
        lov\'{a}sz & - & - & - & 39.6 \\
        focal+lov\'{a}sz & - & - & - & 56.7 \\
        xent+lov\'{a}sz & - & - & - & \textbf{59.1} \\
        \midrule
        xent+lov\'{a}sz & + & - & - & 59.8\\
        xent+lov\'{a}sz & + & + & - & \textbf{63.2} \\
        \midrule
        xent+lov\'{a}sz & + & + & no loss\textsubscript{range} & 62.0 \\
        xent+lov\'{a}sz & + & + & loss\textsubscript{range} & \textbf{65.9} \\
        \bottomrule
    \end{tabular}
    \caption{Ablation study on the loss function, data pre-processing and range-aware loss.}
    \label{table:ablation}
\end{table} 

Table~\ref{table:ablation} shows the ablation study on several design parameters.  For loss function, the combination of cross-entropy loss (xent) and Lov\'{a}sz-softmax loss gets the best results because the Lov\'{a}sz-softmax loss is unstable alone even with its direct optimization on mIoU.  In addition, both the data normalization and augmentation help improve the accuracy by 3.4\%.  Further regularized with the range-aware loss helps optimize the networks to higher accuracy by 3.9\%.

\paragraph{Impact of $\lambda_{range}$ and $\lambda_{lov\acute{a}sz}$ }
\begin{table}[t]
    \centering
    \begin{tabular}{ c  >{\centering\arraybackslash}m{1cm}  >{\centering\arraybackslash}m{1cm}  >{\centering\arraybackslash}m{1cm}  >{\centering\arraybackslash}m{1cm} }
        \toprule
        \multirow{2}{*}{$\lambda_{range}$} & \multicolumn{4}{c}{IoU(\%)} \\
        \cline{2-5}
        & Car & Ped. & Cyclist & Mean \\
        \midrule
        0.00 (baseline) & 74.7 & 64.2 & 47.2 & 62.0 \\
        0.01 & 78.2 & 70.0 & 45.7 & 64.6 \\
        0.05 & 78.1 & 69.3 & 50.2 & 65.9 \\
        0.10 & 77.5 & 70.3 & 55.6 & 67.8 \\
        0.50 & 77.5 & 69.8 & 55.0 & 67.4 \\
        1.00 & 77.8 & 70.7 & 52.3 & 66.9 \\
        \bottomrule
    \end{tabular}
    \caption{Result comparison of different values for $\lambda_{range}$.}
    \label{table:lambda-range}
\end{table}
\begin{table}[t]
    \centering
    \begin{tabular}{ c  >{\centering\arraybackslash}m{1cm}  >{\centering\arraybackslash}m{1cm}  >{\centering\arraybackslash}m{1cm}  >{\centering\arraybackslash}m{1cm} }
        \toprule
        \multirow{2}{*}{$\lambda_{lov\acute{a}sz}$} & \multicolumn{4}{c}{IoU(\%)} \\
        \cline{2-5}
        & Car & Ped. & Cyclist & Mean \\
        \midrule
        0.00 (baseline) & 80.1 & 62.4 & 50.8 & 64.5 \\
        0.01 & 79.8 & 70.7 & 59.0 & 69.8 \\
        0.05 & 79.6 & 70.9 & 57.8 & 69.4 \\
        0.10 & 79.0 & 71.0 & 58.1 & 69.4 \\
        0.50 & 78.5 & 71.1 & 55.2 & 68.3 \\
        1.00 & 77.6 & 70.8 & 52.6 & 67.0 \\
        \bottomrule
    \end{tabular}
    \caption{Result comparison of different values for $\lambda_{lov\acute{a}sz}$.}
    \label{table:lambda-lovasz}
\end{table}
This subsection shows the ablation study of  $\lambda_{range}$ and $\lambda_{lov\acute{a}sz}$.  For $\lambda_{range}$, we use the range-aware LaserNet with the same training strategy to see how $\lambda_{range}$ improves the accuracy.  We train the model with different values of $\lambda_{range}$ with as $\lambda_{lov\acute{a}sz}$ set to $1.0$ .  We can find that the accuracy of the large objects is not affected by the range loss as shown in Table~\ref{table:lambda-range}.  However, the range loss can definitely improve the accuracy of the small objects compared with baseline. The accuracy of the cyclist can be even improved by $8\%$ by setting $\lambda_{range}$ to $0.1$ when compared with no range loss. Similarly, Table~\ref{table:lambda-lovasz} shows the tuning result of $\lambda_{lov\acute{a}sz}$  along with  $\lambda_{range}$ set to $1.0$. The accuracy of the cyclist has been improved by $8.2\%$ by setting $\lambda_{lov\acute{a}sz}$ to $0.01$ when compared with baseline in Table~\ref{table:lambda-lovasz}.

\subsection{Instance-Level Segmentation}
\paragraph{Implementation Details}
For DBSCAN parameters, we choose $minPts$ value as twice the minimum number of points.  Also, we choose $\epsilon$ value as the half of the average object size. After analyzing the data, we can see that the minimum number of points is about 3.5 in Fig.~\ref{fig:laserid}. Thus,we choose 7 for $minPts$  and 0.7 for $\epsilon$. We have tested different distance functions: the original Euclidean distance function, (A), coordinate Z only scaling, (B), and the proposed function as (\ref{eq:distance}), (C).
\paragraph{Evaluation results}
\begin{table}[t]
    \centering
    \begin{tabular}{ c  >{\centering\arraybackslash}m{1cm}  >{\centering\arraybackslash}m{1cm}  >{\centering\arraybackslash}m{1cm}  >{\centering\arraybackslash}m{1cm} }
        \toprule
        \multirow{2}{*}{Distance} & \multicolumn{4}{c}{IoU(\%)} \\
        \cline{2-5}
        & Car & Ped. & Cyclist & Mean \\
        \midrule
        \textit{Semantic} & \textit{77.8} & \textit{70.2} & \textit{53.5} & \textit{67.2} \\
        \midrule
        (A) & 75.1 & 53.7 & 50.5 & 59.8 \\
        (B) & \textbf{75.2} & 54.0 & 49.2 & 59.5 \\
        Ours (C) & 74.6 & \textbf{58.1} & \textbf{52.7} & \textbf{61.8} \\
        \bottomrule
    \end{tabular}
    \caption{Result comparison of instance segmentation.}
    \label{table:kitti3d-inst}
\end{table}

Table~\ref{table:kitti3d-inst} shows the instance-level segmentation evaluation results.  The Z-only scaling distance function (B) is insufficient to segment instance objects, whose performance is almost the same as the original function (A).  In contrast, our proposed weighted distance function (C) gets 2\% of mIoU improvement.  This post-processing step consumes only 16.48ms with standard deviation 24.13ms on i9-7900X.  The visualization results of RangeSeg is shown in Fig.~\ref{fig:instance-vis}. The proposed method can accurately predict the small and far objects when compared with the previous RIU-Net.

\subsection{Result Comparison on the KITTI Raw Data}
\begin{table}[t]
    \centering
    \begin{tabular}{ c  >{\centering\arraybackslash}m{0.5cm}  >{\centering\arraybackslash}m{0.5cm}  >{\centering\arraybackslash}m{0.5cm}  >{\centering\arraybackslash}m{0.8cm}  >{\centering\arraybackslash}m{0.8cm} }
        \toprule
        \multirow{2}{*}{Architecture} & \multicolumn{4}{c}{IoU(\%)} & \multirow{2}{*}{FPS} \\
        \cline{2-5}
        & Car & Ped. & Cyc. & Mean & \\
        \midrule
        SqueezeSeg \cite{squeezeseg} & 64.6 & 21.8 & 25.1 & 37.2 & 122.2 \\
        PointSeg \cite{pointseg} & 67.4 & 19.2 & 32.7 & 39.8 & 106.7 \\
        SqueezeSegV2 \cite{squeezesegv2} & 73.2 & 27.8 & 33.6 & 44.9 & 79.7 \\
        RIU-Net \cite{riunet} & 62.5 & 22.5 & 36.8 & 40.6 & \textbf{249.8} \\
        \midrule
        Ours(LaserNet) & \textbf{75.6} & \textbf{49.5} & \textbf{50.1} & \textbf{58.4} & 109.1 \\
        Ours(ResNet18) & 75.5 & 44.1 & 49.5 & \textbf{56.3} & 89.2 \\
        \bottomrule
    \end{tabular}
    \caption{Result comparison with the related work.}
    \label{table:kittiraw}
\end{table}
Table~\ref{table:kittiraw} shows the result comparison between RangeSeg and other related works on the KITTI raw data.  We follow \cite{squeezeseg} to split the KITTI raw data \cite{kittiraw}.  We choose the $\lambda_{lov\acute{a}sz}$ as 0.05 and $\lambda_{range}$ as 0.5 which is the best setting in the KITTI raw data.  RangeSeg outperforms other state-of-the-art methods.  For more detailed comparisons, RangeSeg gets significant improvements on small objects.  Also, RangeSeg gets 18.6\% improvements of mIoU compared to PointSeg with almost the same inference speed.
\subsection{Real Time Implementation on Nvidia AGX Xavier}
\begin{table}[t]
    \centering
    \begin{tabular}{ c  >{\centering\arraybackslash}m{2cm}  >{\centering\arraybackslash}m{2cm}  }
        \toprule
        Process & Time Avg.(ms) & Time Std.(ms) \\
        \midrule
        Encoding & 4.75 & 0.97  \\
        Model & 13.75 & 0.26  \\
        DBSCAN & 38.22 & 50.92 \\
        \bottomrule
    \end{tabular}
    \caption{Process time of each operation on NVIDIA\textsuperscript{\textregistered} JETSON AGX Xavier.}
    \label{table:xavier-process}
\end{table}
The range-aware LaserNet is implemented on NVIDIA\textsuperscript{\textregistered} JETSON AGX Xavier for embedded system applications.  In our experiment, we use TensorRT FP16 to optimize our framework.  The processing time of data encoding, models and post-processing is summarized in Table~\ref{table:xavier-process}.  It shows that the fps is about 19Hz with TensorRT FP16 optimization, which is much higher than the 10Hz capture frequency of LiDAR sensors in the KITTI dataset.
\subsection{Synthetic Foggy KITTI Dataset}

\begin{table}[t]
    \centering
    \begin{tabular}{ c  >{\centering\arraybackslash}m{0.5cm}  c  >{\centering\arraybackslash}m{0.5cm}  >{\centering\arraybackslash}m{0.5cm}  >{\centering\arraybackslash}m{0.5cm}  >{\centering\arraybackslash}m{0.7cm} }
        \toprule
        \multirow{2}{*}{Model} & \multirow{2}{*}{Aug.} & Test & \multicolumn{4}{c}{IoU(\%)}  \\
        \cline{4-7}
        & & Data & Car & Ped. & Cyc. & Mean \\
        \midrule
        3 channels & - & clear & 77.2 & 70.2 & 56.2 & 67.9 \\
        3 channels & - & 70m & 52.7 & 62.3 & 11.2 & 32.0 \\
        +defog(A) & - & 70m & 60.3  & 37.9 & 22.2 & 40.1 \\
        \midrule
        2 channels & - & clear & 76.4 & 65.3 & 45.5 & 62.4 \\
        +defog(B) & - & 70m & 52.6 & 62.7 & 43.2 & 52.9 \\
        \midrule
        (A)+(B) & - & 70m & 58.2 & 59.7 & 44.5 & 54.1 \\
        \midrule
        \midrule
        \multirow{2}{*}{2 channels} & \multirow{2}{*}{70m} & clear & 72.2  & 66.4 & 47.6 & 62.1 \\
        & & 70m & 67.7 & 66.7 & 47.5 & 60.6 \\
        \midrule
        \multirow{3}{*}{3 channels} & \multirow{3}{*}{70m} & clear & 73.2 & 69.3 & 49.2 & 63.9 \\
        & & 70m & 68.9 & 68.6 & 49.3 & 62.3 \\
        & & 40m & 62.7 & 64.2 & 44.8 & 57.2 \\
        \midrule
        \midrule
        \multirow{3}{*}{3 channels} & \multirow{3}{*}{40m} & clear & 72.3 & 68.8 & 47.2 & 62.8 \\
        & & 40m & 66.8 & 68.0 & 47.4 & 60.7 \\
        & & 70m & 68.3  & 67.2 & 44.0 & 59.8 \\
        \bottomrule
    \end{tabular}
    \caption{Results of the range-aware LaserNet tested on a synthetic foggy data.}
    \label{table:fog}
\end{table}
\begin{figure}[t]
    \centering
    \subfloat[Clear weather.]{
        \includegraphics[height=!,width=0.5\linewidth,keepaspectratio=true]
        {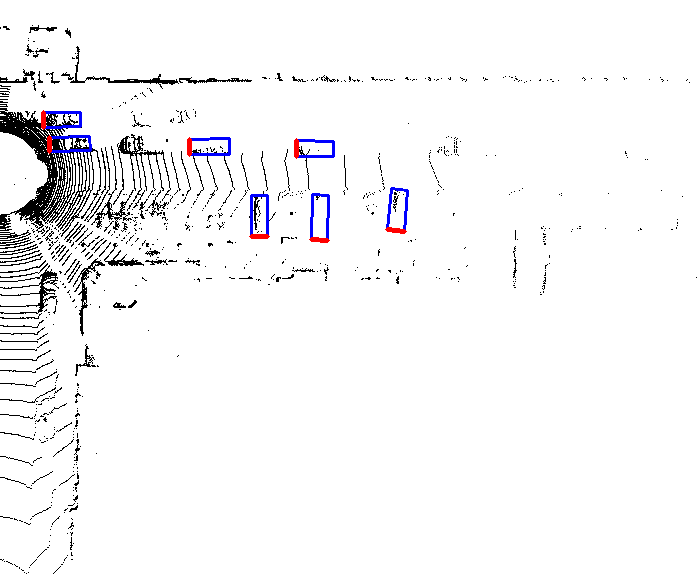}
        \label{fig:bev}
    }
    \subfloat[Visibility at 70 meters.]{
        \includegraphics[height=!,width=0.5\linewidth,keepaspectratio=true]
        {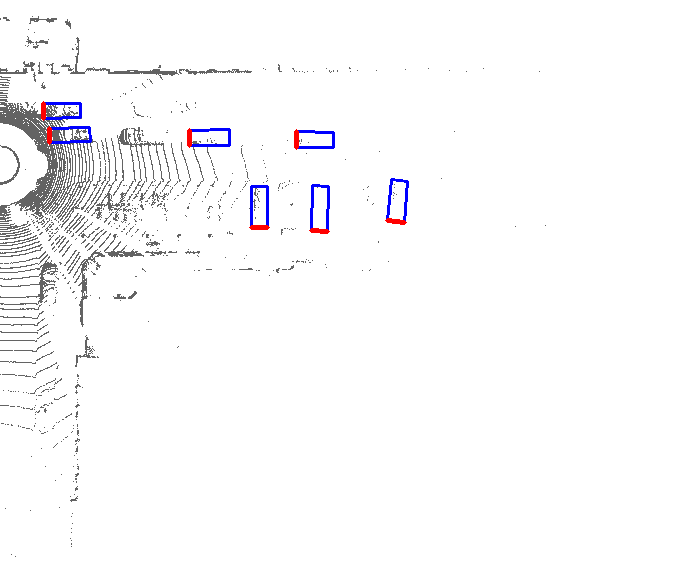}
        \label{fig:bev-fog}
    }
    \caption{The birds' eye views of different weathers.  False measurements will result in an inner circle in the foggy LiDAR point clouds.}
    \label{fig:bev-weather}
\end{figure}
\paragraph{Foggy Dataset Details}
An autonomous driving application should be robust in every environment.  In this experiment, we use our range-aware LaserNet on the synthetic foggy KITTI dataset \cite{foggykitti} with different visibilities.  The visibility is defined as the maximum range that the objects can be visually seen by human.  However, the range of objects can be detected by LiDAR is half of visibility since the point clouds are detected by reflection pulses.  The experiments use visibility at 70m and 40m, which is much more extreme than the worst visibility at 150m in SFSU \cite{foggycity}. The foggy weather will result in false alarm measurements for range within 2 meters mostly, and low intensity reflections for all points. Fig~\ref{fig:bev-weather} shows the birds’ eye view of the LiDAR point clouds in different weathers. The figure shows that there is a blind zone due to car roof mounted LiDAR. The points with wrong range measurements lead to an inner circle as in Fig~\ref{fig:bev-fog}. Therefore, a simple defog method is applied that removes the points shorter than the 2 meters range. 

\paragraph{Evaluation Results}
Table~\ref{table:fog} shows the evaluation results.  When directly testing the model on the foggy data, the accuracy is significantly degraded to 32.0\% of mIoU while the simple defogging (A) only get 8.1\% improvements.  The reason is that the distribution of the intensity channel in the foggy weather is different from that of the clear weather.  Therefore, we train a new model that only contains 2 channels without intensity (denoted as 2 channels).  After defogging on the 2 channels model (B), the accuracy is 52.9\%.  Combining (A) and (B) gets a robust accuracy at 54.1\% of mIoU.

Next, we take foggy data as data augmentation.  The 3 channels model gets higher mIoU with augmentation.  The mIoU is more than 60\% when augmented or tested at 70m or 40m, respectively.  Also, the model trained on visibility at 70m gets 57.2\% of mIoU on visibility at 40m while the model trained on visibility at 40m gets 59.8\% of mIoU on visibility at 70m.  This indicates the LiDAR sensors are robust even in different weather conditions.
\section{Conclusion}
By exploiting the LiDAR data distribution in the autonomous driving application, this paper proposes a range aware instance segmentation network that can achieve high accuracy with a heavy decoder and high speed with a light decoder.  The heavy decoder is applied to the top of the range image where the far and small objects lie in for accurate detection.  The light decoder is applied to the whole range image for low complexity computation.  The proposed weighted distance metric helps segment instances with a simple post-processing.  Compared with previous works, our range-aware framework is simple, efficient, fast and has great applications on autonomous driving in different weathers.  While we have only conducted the experiments on two backbones, further experiments on state-of-the-art models are a potential area for improvements.  Applying this range aware framework to other LiDAR tasks is another interesting future research.

{\small
\bibliographystyle{IEEEtran}
\bibliography{bib}
}

\begin{IEEEbiography}[{\includegraphics[width=1in,height=1.25in,clip,keepaspectratio]{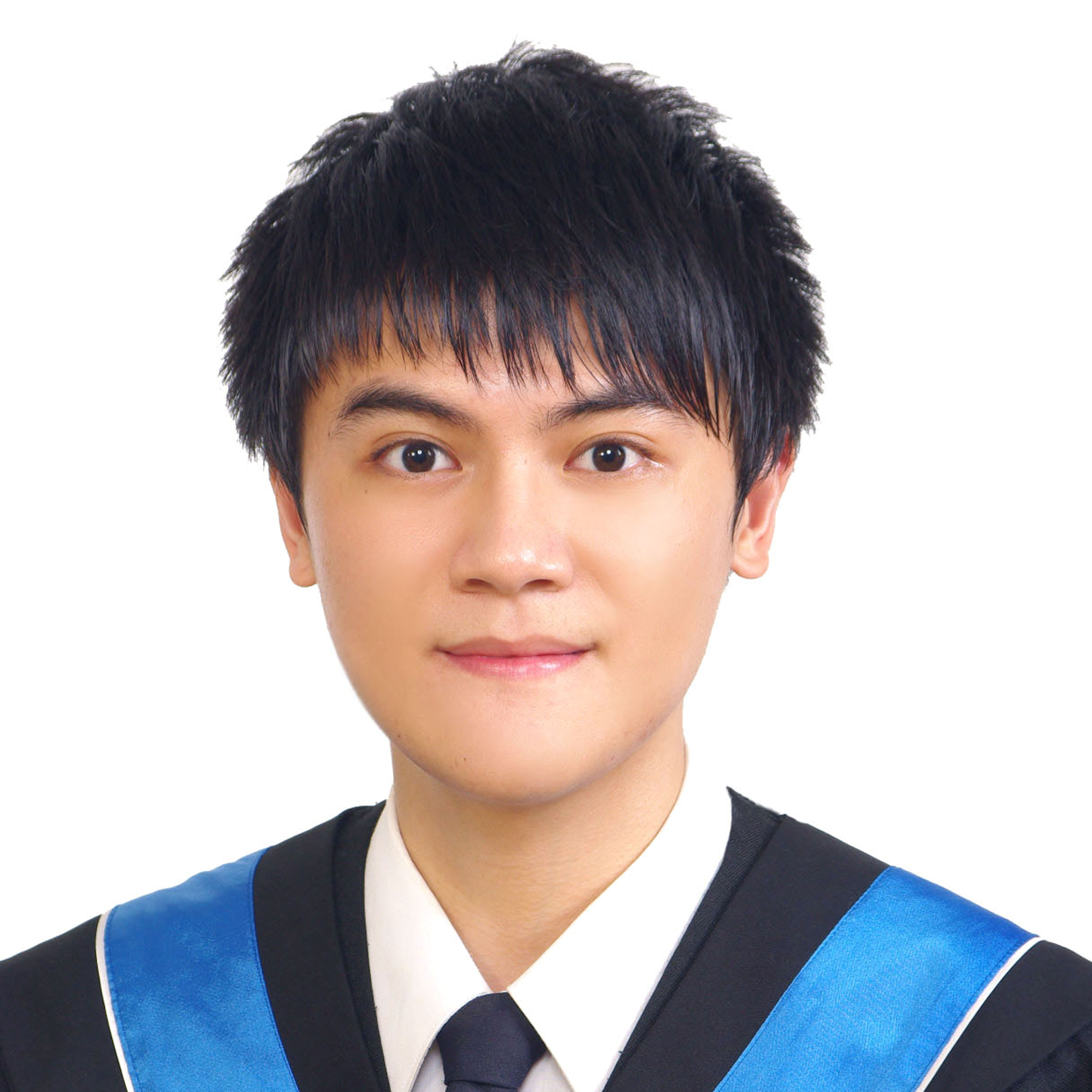}}]{Tzu-Hsuan Chen}
 received the the B.S. and M.S. degrees in electric engineering and electronic engineering from National Chiao-Tung University (NCTU), Hsinchu, Taiwan, in 2017, and 2019 respectively. His current research interests includes VLSI signal processing, and deep learning.

\end{IEEEbiography}

\begin{IEEEbiography}[{\includegraphics[width=1in,height=1.25in,clip,keepaspectratio]{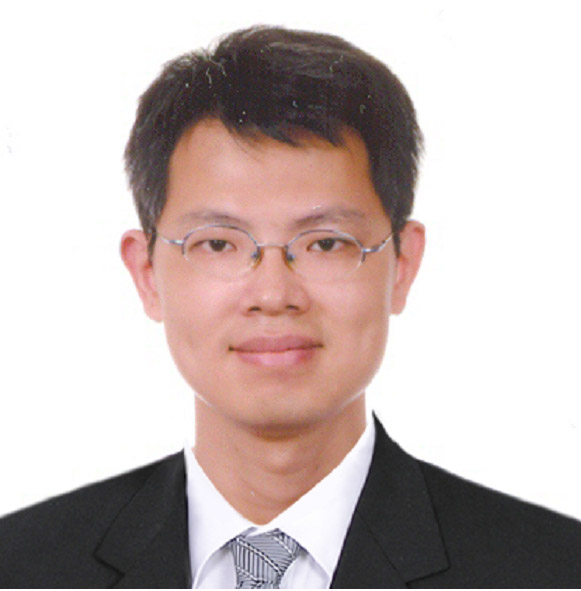}}]{Tian Sheuan Chang}
 (S’93–M’06–SM’07) received the B.S., M.S., and Ph.D. degrees in electronic engineering from National Chiao-Tung University (NCTU), Hsinchu, Taiwan, in 1993, 1995, and 1999, respectively.

	From 2000 to 2004, he was a Deputy Manager with Global Unichip Corporation, Hsinchu, Taiwan. In 2004, he joined the Department of Electronics Engineering, NCTU, where he is currently a Professor. In 2009, he was a visiting scholar in IMEC, Belgium. His current research interests include system-on-a-chip design, VLSI signal processing, and computer architecture.

\end{IEEEbiography}
\end{document}